\documentclass[11pt]{article}

\usepackage[final]{acl}

\usepackage{amsmath}
\usepackage{booktabs}
\usepackage{multirow}
\usepackage{amssymb}
\usepackage[table]{xcolor}
\usepackage{pifont}
\newcommand{\gain}[1]{\textcolor{teal}{\scriptsize (+#1)}}
\newcommand{\cmark}{\textcolor{teal}{\ding{51}}}
\newcommand{\xmark}{\textcolor{red!70}{\ding{55}}}
\newcommand{\umark}{\textcolor{gray}{--}}
\newcommand{\symboletongyi}{\raisebox{0pt}{~\includegraphics[scale=0.012]{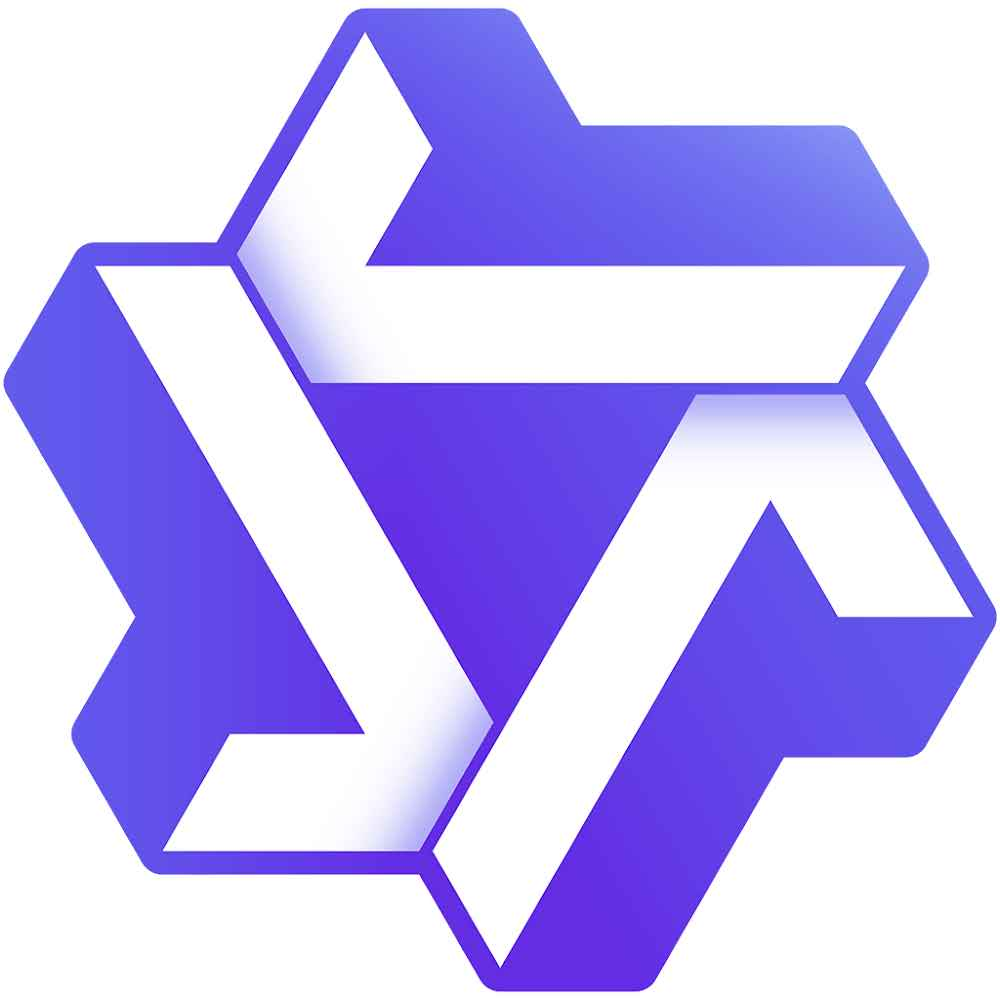}}~}
\usepackage{times}
\usepackage{latexsym}
\usepackage{hyperref}
\usepackage[T1]{fontenc}
\usepackage[utf8]{inputenc}
\usepackage{microtype}
\usepackage{inconsolata}

\usepackage{graphicx}

\title{EvoTrainer: Co-Evolving LLM Policies and Training Harnesses for Autonomous Agentic Reinforcement Learning}

\author{
  \textbf{Guhong Chen\textsuperscript{1}},
  \textbf{Yingcheng Shi\textsuperscript{2}},
  \textbf{Yongbin Li\textsuperscript{2,$\dagger$}},
  \textbf{Binhua Li\textsuperscript{2}},
  \textbf{Xander Xu\textsuperscript{3}},
\\
  \textbf{Hu Wei\textsuperscript{3}},
  \textbf{Shiwen Ni\textsuperscript{4}},
  \textbf{Min Yang\textsuperscript{1,4,$\dagger$}},
  \textbf{Jieping Ye\textsuperscript{2}}
\\
\\
  \textsuperscript{1}Shenzhen Institutes of Advanced Technology, Chinese Academy of Sciences
\\
  \textsuperscript{2}Tongyi Lab\symboletongyi, Alibaba Group \quad
  \textsuperscript{3}Alibaba Group \quad
  \textsuperscript{4}SUAT
}

\begin{document}
\maketitle

\begingroup
  \renewcommand\thefootnote{$\dagger$}
  \footnotetext{
    Corresponding authors: 
    \href{mailto:shuide.lyb@alibaba-inc.com}{shuide.lyb@alibaba-inc.com}, 
    \href{mailto:min.yang@siat.ac.cn}{min.yang@siat.ac.cn}. \\
    \hspace*{1.8em}\small{\mbox{Code available at:}\,\url{https://github.com/AlibabaResearch/DAMO-ConvAI/tree/main/EvoTrainer}}
  }
\endgroup

\begin{abstract}
Autonomous LLM training is often framed as recipe search, which leaves the training harness largely static. This limitation sharpens in agentic RL, where shifting bottlenecks and scalar rewards mask diverse failure modes. We introduce \textbf{EvoTrainer}, an autonomous training framework that co-evolves LLM policies and training-side harnesses through empirical feedback: it diagnoses rollout-level evidence, revises diagnostics, backtests interventions, and accumulates reusable skills. Evaluated on mathematical reasoning, competitive-programming code generation, and repository-level software engineering, EvoTrainer matches or exceeds the human-engineered RL references under the same data, codebase, and evaluation protocol, with the largest gain on long-horizon agentic SWE. Trajectory analyses show that retained strategies diverge across domains, evolving diagnostics prevent invalid high-scoring branches from being promoted, and reusable skills shape later search. Autonomous LLM RL should move beyond recipe search toward joint evolution of policies and the training harnesses that interpret them.

\end{abstract}

\section{Introduction}

AI systems are beginning to participate in model development by editing code, launching experiments, inspecting outcomes, and proposing new training versions \citep{karpathy2026autoresearch,lu2024aiscientist,yamada2025aiscientistv2,ning2026specialistautoresearch,jeddi2026gear}. These systems suggest that future model improvement may depend not only on human-designed training recipes, but also on agents that can iteratively revise them through empirical feedback.

\begin{figure}[t]
\centering
\includegraphics[width=\columnwidth]{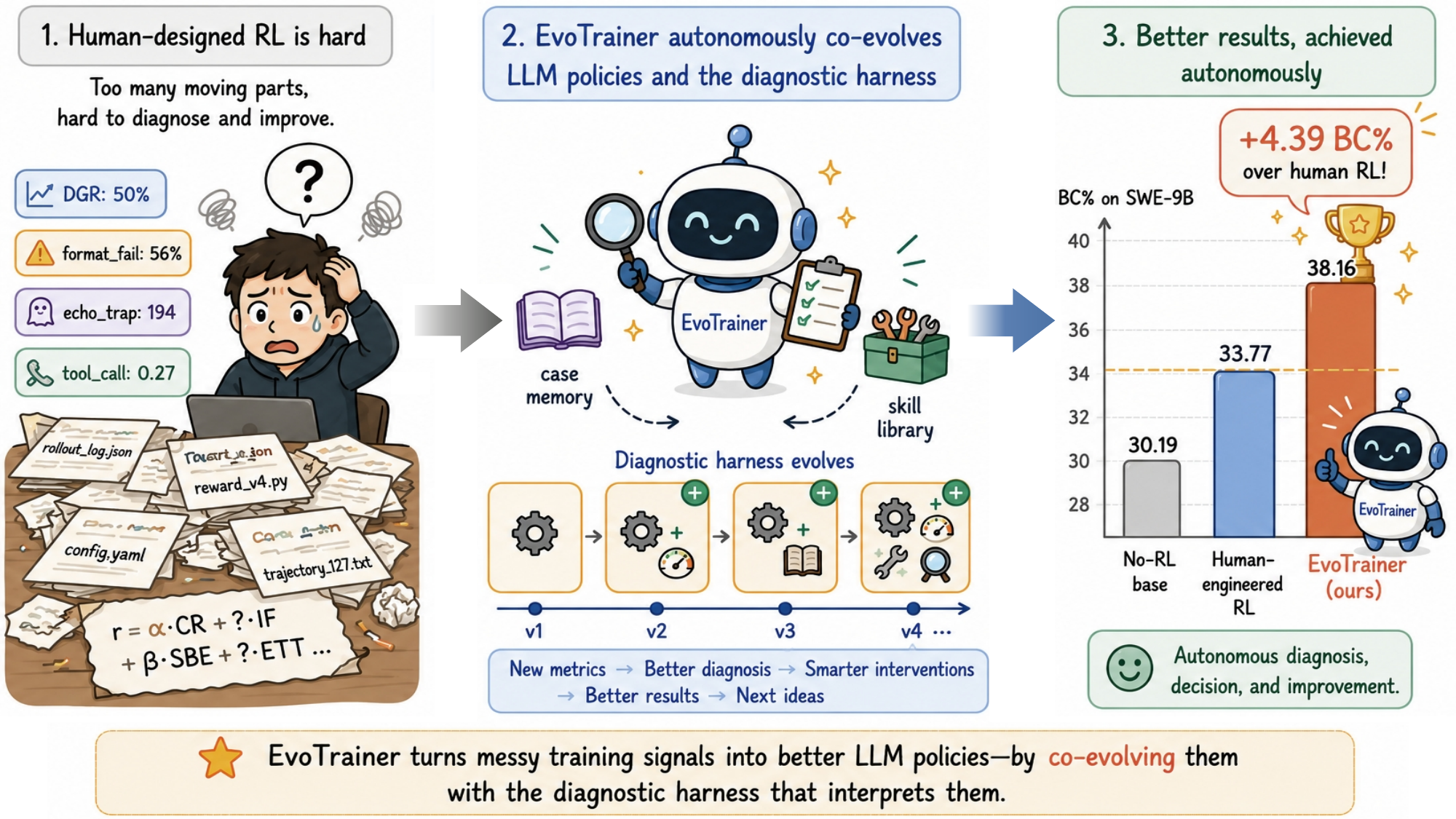}
\caption{
Overview of EvoTrainer: an autonomous training framework that co-evolves LLM policies and training-side diagnostic harnesses, exceeding the human-engineered RL baseline on SWE-9B by \textbf{+4.39 BC\%}.
}
\vspace{-0.4cm}
\label{fig:teaser}
\end{figure}

However, most autonomous experimentation systems still keep the decision infrastructure around training largely fixed: they search over candidate recipes yet rely on the same diagnostic views, memory, and intervention logic to interpret each new result. This is limiting in complex reinforcement learning, where the dominant bottleneck may shift from reward sparsity to behavior collapse, from evaluation artifacts to low-information rollout groups, or from recipe selection to the need for reusable diagnostic tools. Scalar validation scores are one visible failure mode; the broader problem is that the evidence and procedures needed to guide training may themselves need to evolve.

This challenge is especially pronounced in agentic RL. In such settings, a model may search files, invoke tools, edit code, execute tests, inspect error messages, and submit a final solution only after many turns. The resulting training process is difficult to steer with a fixed diagnostic template: successful scores may hide reward leakage or unhealthy behaviors, failed branches may reveal valuable negative evidence, and later versions may require analyses that were unnecessary at earlier stages. The diagnostics needed to interpret training outcomes therefore often change across versions and are difficult to specify fully in advance.

This paper studies the training system itself as an object of improvement. We use the term trainer to denote the decision-making system that observes completed versions, analyzes rollout evidence, proposes interventions, updates diagnostic infrastructure, and determines what should be tested next. The policy improves within a training run; the trainer improves across training runs by accumulating evidence, revising its harness, and reusing operational skills.

We propose \textbf{EvoTrainer} (Figure~\ref{fig:teaser}), an autonomous training framework that co-evolves LLM policies and training-side diagnostic harnesses through two coupled processes: policy self-evolution, where runnable training versions are generated, compared, pruned, promoted, and merged through controlled interventions; and trainer self-reflection, where the training-side harness evolves when existing metrics, analyzers, backtests, or search procedures are insufficient. A persistent memory and reusable skill library let later iterations retrieve failed-branch lessons, diagnostic scripts, and previously validated mechanisms. The trainer agent autonomously runs this loop by constructing versions, diagnosing outcomes, revising the harness, and proposing interventions, while humans bootstrap the workspace and approve costly or consequential execution (Section~\ref{subsec:autonomy-scope}).

Table~\ref{tab:system-comparison} situates EvoTrainer among representative autonomous experimentation systems. AutoResearch and Bilevel Autoresearch target training-recipe optimization on GPT pretraining benchmarks \citep{karpathy2026autoresearch,qu2026bilevelautoresearch}. GEAR introduces population-style search over agentic code \citep{jeddi2026gear}; Meta-Harness and AHE optimize inference-side harnesses for LLM applications \citep{lee2026metaharness,lin2026ahe}. To our knowledge, EvoTrainer is the first autonomous training framework in agentic LLM RL to treat the training-side diagnostic harness itself as an evolving object.

\begin{table*}[t]
\centering
\footnotesize
\setlength{\tabcolsep}{3.5pt}
\begin{tabular}{lccccc}
\toprule
\textbf{Method}
& \shortstack{\textbf{Autonomous}\\\textbf{Experiment Loop}}
& \shortstack{\textbf{Inference-Side}\\\textbf{Harness}}
& \shortstack{\textbf{Training-Side}\\\textbf{Harness}}
& \shortstack{\textbf{Reusable}\\\textbf{Skill Library}}
& \shortstack{\textbf{Agentic}\\\textbf{RL}} \\
\midrule
AutoResearch~\citep{karpathy2026autoresearch}
& \cmark & \xmark & \xmark & \xmark & \xmark \\
Bilevel Autoresearch~\citep{qu2026bilevelautoresearch}
& \cmark & \xmark & \umark & \xmark & \xmark \\
GEAR~\citep{jeddi2026gear}
& \cmark & \xmark & \xmark & \xmark & \xmark \\
Meta-Harness~\citep{lee2026metaharness}
& \cmark & \cmark & \xmark & \xmark & \xmark \\
AHE~\citep{lin2026ahe}
& \cmark & \cmark & \xmark & \cmark & \xmark \\
EvoTrainer
& \cmark & \xmark & \cmark & \cmark & \cmark \\
\bottomrule
\end{tabular}
\caption{
Capability comparison with representative autonomous experimentation systems.
Inference-side harness evolution optimizes scaffolding around the model at inference time (context, tools, memory); training-side harness evolution revises the diagnostic infrastructure that interprets training-time outcomes.
Symbols: \cmark{} = capability supported; \xmark{} = not present per the cited paper; \umark{} = unclear from the cited paper.
}
\vspace{-0.3cm}
\label{tab:system-comparison}
\end{table*}

We evaluate EvoTrainer on mathematical reasoning, competitive-programming code generation, and repo-level software engineering. EvoTrainer consistently improves over the no-RL base in every domain and matches or exceeds the human-engineered RL references developed under the same data, codebase, and evaluation protocol, with the largest gain on SWE-9B: 38.16 BC\% versus 30.19 for no-RL and 33.77 for the human-engineered RL baseline. Component-level analyses further show that retained strategies diverge across domains, the evolving harness rejects invalid high-scoring branches, and retained skills alter later search, providing process-level evidence beyond score-driven iteration.

Our contributions are threefold: (i) we formulate autonomous model training as cross-version trainer improvement, where adaptation targets both the model recipe and the decision infrastructure that interprets outcomes; (ii) we introduce EvoTrainer, a dual-evolution framework that jointly develops policy versions and a training-side diagnostic harness through signal diagnosis, harness revision, persistent memory, and reusable skills; and (iii) we evaluate EvoTrainer across Math, Coding, and SWE, showing that it matches or exceeds human-engineered RL references and providing process-level evidence beyond score-dominant iteration.

\section{Related Work}
\label{sec:related}

\subsection{Autonomous Research and Self-Evolving Experimentation}

Recent work automates increasing portions of scientific discovery and model development. AutoResearch demonstrates a propose--train--evaluate loop on GPT pretraining benchmarks, while Bilevel Autoresearch meta-optimizes the inner research loop~\citep{karpathy2026autoresearch,qu2026bilevelautoresearch}. Specialist-agent frameworks cast training-recipe optimization as auditable trajectories with failure-aware feedback~\citep{ning2026specialistautoresearch}. The AI Scientist line extends this toward end-to-end scientific discovery~\citep{lu2024aiscientist,yamada2025aiscientistv2}. GEAR introduces population-based search over agentic code agents~\citep{jeddi2026gear}. A related line of self-improving systems evolves code, algorithms, or training curricula through empirical feedback~\citep{zhang2025dgm,novikov2025alphaevolve,huang2025rzero,yu2026easyrl,tao2024selfevolution}. EvoTrainer extends this line to agentic LLM RL training, where the trainer must additionally evolve a training-side diagnostic harness to interpret rollout-level evidence and steer interventions across versions.

\subsection{Harness and Infrastructure Optimization for LLM Systems}

System performance depends on infrastructure surrounding the model, not only on model weights. Meta-Harness searches over harness code and shows automatically discovered task-side harnesses can outperform hand-engineered designs~\citep{lee2026metaharness}. AHE evolves coding-agent harnesses through observability-driven trajectory analysis~\citep{lin2026ahe}; TDScaling uses diversity-sensitive diagnostics to steer trajectory synthesis for code agents~\citep{chen2026tdscaling}. These works target inference-time scaffolding around an LLM rather than the training process itself. EvoTrainer operates on a different layer: a training-side diagnostic harness that accumulates metrics, analyzers, backtests, retrieved evidence, and reusable skills to interpret policy-version outcomes during RL training. The signal substrate it targets---reward distributions, group variance, behavioral trajectories, dead-group ratios, and cross-version intervention evidence---is not primarily addressed by inference-time harness work. Adapting those systems would require reconstructing their search and evaluation logic around training-time RL artifacts.

\subsection{Task-Adaptive RL Design for Verifiable and Agentic Training}

Recent RL methods for language models introduce specialized mechanisms: group-relative updates in GRPO~\citep{shao2024deepseekmath}, Clip-Higher and dynamic sampling in DAPO~\citep{yu2025dapo}, and sequence-level optimization in GSPO~\citep{zheng2025gspo}. Complementary work in verifiable-reward training documents optimization bias, diversity-aware reward design, adaptive guidance, and verifiable-environment construction~\citep{liu2025drgrpo,chen2025dragrpo,liu2025ghpo,zeng2025rlve,huang2026rlvr}, indicating RL recipes are highly sensitive to task structure, reward granularity, data regime, and model scale.

This sensitivity sharpens in agentic RL: RAGEN identifies Echo Trap and motivates trajectory-level stabilization~\citep{wang2025ragen}; RAGEN-2 proposes SNR-aware variance filtering~\citep{wang2026ragen2}. Long-horizon and tool-using agent studies document strong sensitivity to reward shaping and environment stability~\citep{chen2025longhorizonagents,wu2026demystifying}. EvoTrainer addresses this adaptation layer by letting the trainer diagnose version-specific failures, retrieve or revise candidate mechanisms, and retain only interventions supported by cross-version evidence.

\section{EvoTrainer: Co-Evolving LLM Policies and Training Harnesses}
\label{sec:trace}

\begin{figure*}[t]
    \centering
    \includegraphics[width=0.98\linewidth]{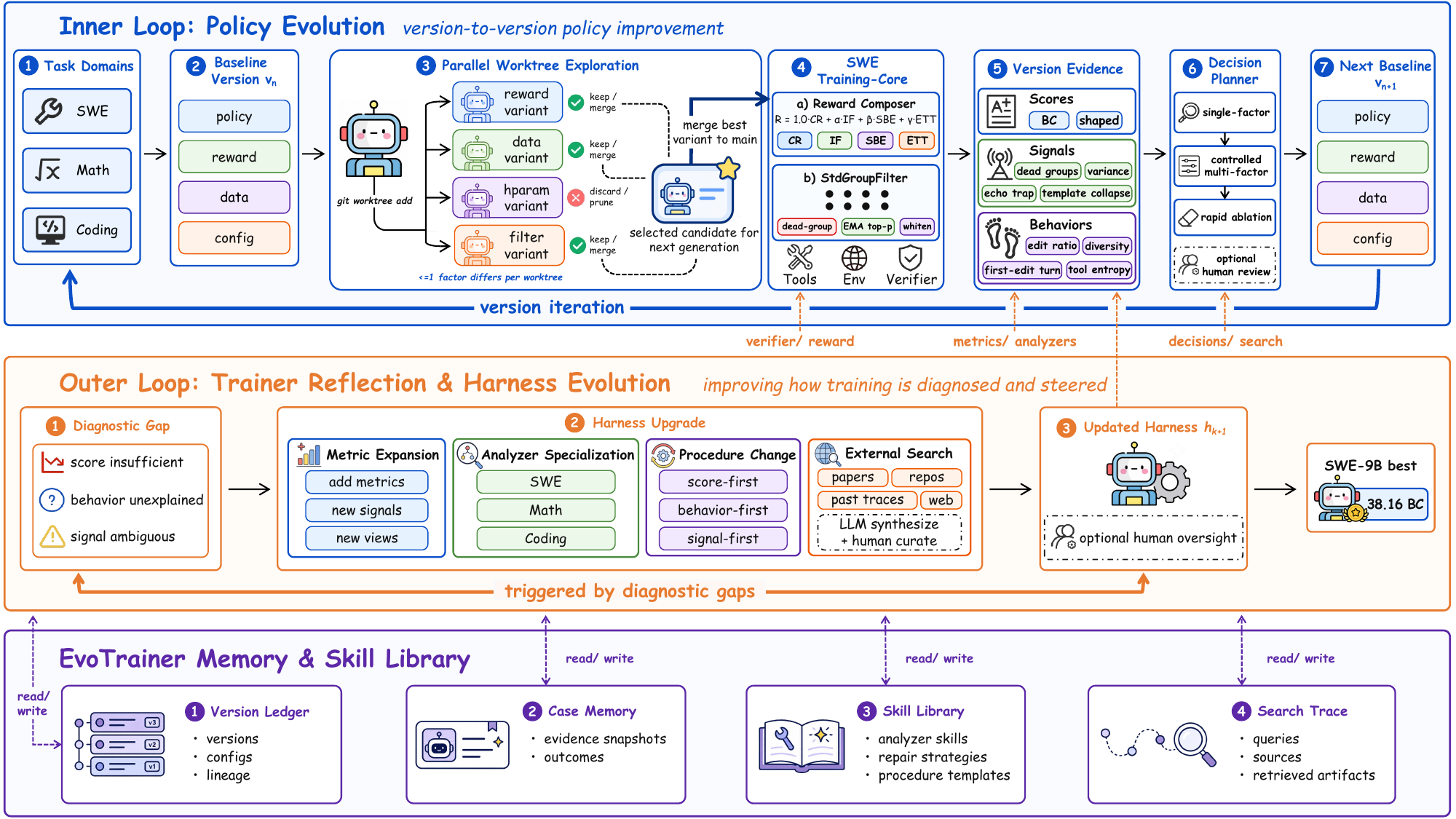}
    \caption{
    Overview of EvoTrainer.
    The upper loop evolves policy versions through controlled exploration, training, evidence collection, and intervention planning; the middle loop evolves the training-side diagnostic harness; and the bottom layer stores persistent memory and reusable skills.
    The training-core panel illustrates the SWE instantiation.
    }
     \vspace{-0.3cm}
    \label{fig:trace-overview}
\end{figure*}

\subsection{Versioned Autonomous Training}
\label{subsec:versioned-training}

An autonomous trainer must do more than execute training jobs: given a completed version, it must determine what the outcome means, which intervention was responsible, what failure mode emerged, and which direction should be tested next. We therefore formulate autonomous training as a sequence of evidence-conditioned version transitions in which both policy versions and the training-side diagnostic harness evolve jointly.

Let \(v_0, v_1, \ldots, v_n\) denote the evolving policy versions. Each version \(v_i\) produces artifacts \(\mathcal{A}_i=\{\mathrm{metrics}_i,\mathrm{rollouts}_i,\mathrm{configs}_i,\mathrm{logs}_i,\mathrm{diffs}_i\}\), which are interpreted by the current training harness \(h_i\). We summarize a completed training state as \(\mathcal{T}_i=(v_i,h_i,\mathcal{A}_i,d_i,\Delta_i,\omega_i)\), where \(d_i\) is the diagnosis of the current version, \(\Delta_i\) is the proposed intervention, and \(\omega_i\) is the observed outcome. The outcome may indicate improvement, regression, mixed evidence, or insufficient evidence.

This formulation makes the version transition the unit of autonomous improvement: a score-improving version may still expose a fragile reward design, a regressed branch may reveal a harmful intervention, and a mixed outcome may show one bottleneck resolved while another becomes visible. EvoTrainer preserves these distinctions instead of collapsing each run into a binary keep-or-reject event. The two layers are then developed separately: policy self-evolution (Section~\ref{subsec:model-self-evolution}) and training-side harness evolution (Section~\ref{subsec:harness-evolution}).

\subsection{Policy Self-Evolution through Version-Controlled Exploration}
\label{subsec:model-self-evolution}

The trainer agent first constructs a runnable training version with executable launch scripts, reward wiring, configuration files, and evaluation hooks; subsequent versions evolve through controlled exploration.

At version \(v_i\), EvoTrainer creates a candidate set \(\mathcal{B}_i=\{b_i^{(1)},b_i^{(2)},\ldots,b_i^{(K)}\}\), where each branch applies an intervention \(\Delta_i^{(k)}\) to the current baseline:
\[
b_i^{(k)} = v_i \oplus \Delta_i^{(k)}.
\]
Here, \(\oplus\) denotes a versioned modification to the current training recipe. A candidate change may target rewards, filtering, data selection, rollout settings, optimization choices, or tool-use behavior.

EvoTrainer defaults to single-factor interventions for clean attribution; bundled changes are admitted only when each component has prior independent support or when the interaction itself is being tested, with the rationale recorded in the version ledger and ambiguous outcomes revisited through targeted ablations or backtests.

Branches are materialized as isolated worktrees and can be trained in parallel when resources permit. After training, the trainer compares the resulting evidence and recommends whether to keep, prune, revert, or merge each branch. A promoted branch becomes the next baseline \(v_{i+1}\), while failed or ambiguous branches remain useful as negative evidence. This version-control discipline makes autonomous experimentation auditable through an explicit lineage of explored interventions and their consequences.

\subsection{Trainer Reflection and Harness Evolution}
\label{subsec:harness-evolution}

A fixed diagnostic harness is often insufficient for long-horizon autonomous training. Early versions may be analyzable through coarse validation trends, whereas later versions may require new behavioral metrics, reward audits, group-variance statistics, failure taxonomies, retrieval procedures, or backtesting scripts. EvoTrainer therefore treats the training harness itself as an evolving object.

We organize the harness around four diagnostic layers: \emph{score} (validation metrics, benchmark-level improvements), \emph{signal} (reward variance, dead-group ratio, component-level reward adoption), \emph{behavior} (tool-use patterns, search/edit/test structure, trajectory length, degeneration modes, environment-side anomalies), and \emph{version} (cross-branch promotion, rejection, and retention decisions). These layers are complementary: reward statistics from a rollout provide signal evidence, while action patterns provide behavior evidence.

A diagnostic gap appears when existing evidence cannot support one of three functions: explaining a version outcome, distinguishing competing hypotheses, or selecting a defensible next intervention. EvoTrainer upgrades the harness along four axes: metric expansion (e.g., dead-group ratio, rollout diversity, first-edit turn); analyzer specialization (per-domain analyzer routines); procedure revision (e.g., shifting from score-first to behavior-first analysis); and external evidence retrieval (querying papers, repositories, and prior traces to synthesize candidate interventions or harness updates).

Training-signal revisions surface from harness reflection but are executed via the policy-evolution mechanism of Section~\ref{subsec:model-self-evolution}; Section~\ref{subsec:framework-evidence} examines its empirical effect.

\subsection{Persistent Memory and Reusable Skill Library}
\label{subsec:memory}

EvoTrainer maintains a persistent memory layer \(\mathcal{M}=\{\mathcal{L}_{\mathrm{ver}},\mathcal{C}_{\mathrm{case}},\mathcal{S}_{\mathrm{skill}},\mathcal{S}_{\mathrm{search}}\}\). \(\mathcal{L}_{\mathrm{ver}}\) stores model lineage, configuration diffs, Git diffs, and keep/prune/merge decisions. \(\mathcal{C}_{\mathrm{case}}\) stores recurring patterns: score-and-behavior divergence, low-information rollout groups, or interventions that repeatedly fail under a particular domain structure. \(\mathcal{S}_{\mathrm{skill}}\) stores reusable analyzer skills, repair strategies, and procedure templates that, once validated, later versions can call, extend, or adapt. \(\mathcal{S}_{\mathrm{search}}\) stores retrieval queries, external sources, distilled insights, and adopted patches or procedures.

Memory thus turns EvoTrainer from isolated experiments into a cumulative process: a SWE filtering utility is later retrieved in Math and Coding when zero-variance groups recur, showing operational rather than archival reuse (Section~\ref{subsec:framework-evidence}).

\subsection{Trainer-Agent Implementation and Autonomy Scope}
\label{subsec:autonomy-scope}

In our experiments, the trainer is instantiated with Claude Sonnet 4.6~\citep{anthropic2026claudesonnet46}; the workflow is replaceable at the trainer-model interface, provided the trainer has access to repository files, shell execution, experiment artifacts, and retrieval utilities. After humans bootstrap the task, benchmark, data, compute environment, and initial playbooks, the trainer agent autonomously runs the diagnostic loop and recommends interventions, while costly or consequential actions remain human-gated (Table~\ref{tab:autonomy-scope}). This separation makes the training cognition loop autonomous while keeping costly or irreversible execution decisions under human control.

\begin{table}[t]
\centering
\footnotesize
\setlength{\tabcolsep}{3.5pt}
\renewcommand{\arraystretch}{1.15}
\begin{tabular}{lcc}
\toprule
\rowcolor{gray!18}
\textbf{Action} & \textbf{Trainer} & \textbf{Human} \\
\midrule
\rowcolor{blue!8}
\multicolumn{3}{l}{\textit{Setup}} \\
Workspace bootstrapping &  & \cmark \\
First runnable training version & \cmark &  \\
\midrule
\rowcolor{teal!10}
\multicolumn{3}{l}{\textit{Autonomous diagnostic loop}} \\
Artifact analysis and diagnosis & \cmark &  \\
Harness updates and skill creation & \cmark &  \\
External search and backtesting & \cmark &  \\
Intervention proposal & \cmark &  \\
\midrule
\rowcolor{orange!10}
\multicolumn{3}{l}{\textit{Gated execution}} \\
Policy training-run launch & drafts & approves \\
Version promotion / pruning & recommends & confirms \\
\bottomrule
\end{tabular}
\caption{
Autonomy scope in EvoTrainer: humans set up the workspace and gate costly or consequential execution, while the trainer agent performs the core iterative diagnostic loop.
}
\vspace{-0.3cm}
\label{tab:autonomy-scope}
\end{table}

\subsection{SWE Training-Core Instantiation}
\label{subsec:swe-training-core}

The final SWE trajectory uses a GRPO-style training core~\citep{shao2024deepseekmath} with asymmetric Clip-Higher~\citep{yu2025dapo} bounds and weak KL regularization to a fixed reference policy (full objective in Appendix~\ref{app:training-core}). For each prompt \(q\), the old policy samples \(G\) trajectories \(\{o_1, \ldots, o_G\}\), each receiving a scalar reward \(r_i\). The group-relative advantage is
\[
\begin{aligned}
A_i &= \frac{r_i - \mu_g}{\sigma_g + \epsilon}, \\
\mu_g &= \frac{1}{G}\sum_{j=1}^{G} r_j, \quad \sigma_g = \mathrm{std}(r_1,\ldots,r_G).
\end{aligned}
\]
We broadcast \(A_i\) to all tokens of trajectory \(o_i\), avoiding a learned value model but making group variance critical: when \(\sigma_g=0\), the group yields no useful relative learning signal.

The final SWE reward design combines task correctness and behavior-sensitive components:
\[
r_i
=
1.0 \cdot \mathrm{CR}_i
+
0.1 \cdot \mathrm{IF}_i
+
0.1 \cdot \mathrm{SBE}_i
+
0.15 \cdot \mathrm{ETT}_i.
\]
Here, \(\mathrm{CR}\) is hidden-test correctness, \(\mathrm{IF}\) is instruction-following reward from a frozen judge, \(\mathrm{SBE}\) rewards search-before-edit behavior, and \(\mathrm{ETT}\) rewards edit-then-test behavior. These components are not introduced as a fixed handcrafted recipe; they emerge through the versioned evolution process analyzed in Section~\ref{subsec:framework-evidence} and Appendix~\ref{app:swe-trajectory}. The asymmetric SBE/ETT weighting reflects a trainer-side decision: an earlier symmetric variant (\(0.1\cdot\mathrm{SBE}+0.1\cdot\mathrm{ETT}\)) produced reward ties across a substantial fraction of trajectories, collapsing within-group variance and neutralizing the procedural reward signal.

Before advantage normalization, EvoTrainer applies variance-aware group filtering. Groups with zero reward variance are removed as dead groups, and an adaptive filtering rule suppresses low-information groups before policy updates. This mechanism is especially important in SWE, where sparse hidden-test outcomes can otherwise produce large fractions of non-informative groups. Math and Coding use the same GRPO core but with domain-specific reward structures that emerge through the evolution process described in Section~\ref{subsec:domain-evolution}.

\section{Experiments}
\label{sec:experiments}

We evaluate EvoTrainer on three domains with distinct training and diagnostic demands: mathematical reasoning, competitive-programming code generation, and repo-level software engineering. Math and Coding are single-turn generation settings, whereas SWE requires long-horizon agent--tool interaction in an executable environment. Together, these domains test whether EvoTrainer remains useful across training regimes with increasingly complex decision requirements, from single-turn verifiable tasks to long-horizon agentic interaction.

\subsection{Experimental Setup}
\label{subsec:experimental-setup}

\paragraph{Tasks and Data.}
For Math, we train on 6,429 problems from BigMath-Hard~\citep{albalak2025bigmath} (one example overlapping AIME 2024 P5 removed) and evaluate on 78 competition problems: 30 AIME 2024, 30 AIME 2025, 18 CNMO 2024. For Coding, we train on 11,897 verified problems from TACO-verified~\citep{li2023taco} and evaluate on 175 held-out problems from a recent LiveCodeBench-v6 subset of AtCoder Beginner Contest tasks, with earlier-released problems excluded to avoid contamination~\citep{jain2024livecodebench}; training and validation are disjoint, yielding an out-of-distribution protocol. For SWE, we train on 8,622 instances from the swe-rebench-v6 train split and evaluate on 77 held-out Python instances from the corresponding test split~\citep{badertdinov2025swerebench}; each instance runs in a Docker environment judged by hidden fail-to-pass tests, with the tool scaffold, interaction protocol, and evaluation harness frozen across compared methods.

\paragraph{Evaluation Protocol.}
All domains use Avg@8 (mean over 8 independent rollouts per item) with a single random seed (seed \(42\)). Math correctness is determined by a frozen Qwen3.5-4B judge~\citep{qwen2026qwen35_4b}; Coding correctness by stdin/stdout execution; SWE by BC\%, the proportion of valid executions passing hidden fail-to-pass tests (excluding infrastructure-level failures that do not reflect model behavior).

\paragraph{Baselines.}
We compare EvoTrainer against three groups of references. The first contains the no-RL base model, the strongest human-engineered RL configuration (developed by the same research team under the same codebase, model family, data, and evaluation protocol), and AutoResearch~\citep{karpathy2026autoresearch} as a score-only autonomous iteration reference. The second contains representative algorithmic baselines instantiated within the same training stack: GRPO and GSPO variants~\citep{shao2024deepseekmath,zheng2025gspo}, Clip-Higher stabilization from DAPO~\citep{yu2025dapo}, RAGEN-style filtering variants~\citep{wang2025ragen,wang2026ragen2}, and a group-variance filtering baseline. The third reports the retained EvoTrainer version. Appendix~\ref{app:human-budget} reports search-budget, training-compute, and trainer-agent inference accounting for transparency. The other autonomous experimentation frameworks listed in Table~\ref{tab:system-comparison} target search spaces and evaluators that differ from LLM RL training and are not directly comparable.

\subsection{Does EvoTrainer Improve Training Across Domains?}
\label{subsec:main-results}

\begin{table*}[t]
\centering
\small
\setlength{\tabcolsep}{4.0pt}
\begin{tabular}{lcccccc}
\toprule
\multirow{2}{*}{\textbf{Method}}
& \multicolumn{2}{c}{\textbf{SWE Avg@8 BC\%}}
& \multicolumn{3}{c}{\textbf{Math Avg@8}}
& \multirow{2}{*}{\textbf{Coding Avg@8}} \\
\cmidrule(lr){2-3} \cmidrule(lr){4-6}
& \textbf{4B} & \textbf{9B}
& \textbf{AIME 2024} & \textbf{AIME 2025} & \textbf{CNMO 2024}
& \\
\midrule
\multicolumn{7}{l}{\textit{Reference Configurations}} \\
Base model (no RL)
& 24.68 & 30.19
& 77.50 & 67.50 & 75.00
& 46.71 \\
Human-engineered RL
& 31.17 & 33.77
& 80.83 & 71.67 & 77.78
& 50.71 \\
AutoResearch~\citep{karpathy2026autoresearch}
& 28.41 & 33.33
& 78.33 & 70.42 & 78.47
& 45.51 \\
\midrule
\multicolumn{7}{l}{\textit{Algorithmic Baselines}} \\
GSPO + Clip-Higher
& 24.84 & 31.97
& 79.17 & 70.00 & 76.39
& 47.86 \\
GRPO + Clip-Higher
& 25.49 & 32.79
& 80.00 & 69.58 & 79.17
& 48.57 \\
RAGEN v1 Filtering
& 29.87 & 32.89
& 80.42 & 70.83 & 77.78
& 49.04 \\
RAGEN v2 SNR Filtering
& 30.03 & 35.74
& 81.67 & 72.50 & 80.56
& 49.86 \\
Group-Variance Filtering
& 29.06 & 32.31
& 78.75 & 68.75 & 76.39
& 48.21 \\
\midrule
\textbf{EvoTrainer}
& \textbf{31.49} \gain{6.81}
& \textbf{38.16} \gain{7.97}
& \textbf{84.17} \gain{6.67}
& \textbf{73.33} \gain{5.83}
& \textbf{81.94} \gain{6.94}
& \textbf{51.29} \gain{4.58} \\
\bottomrule
\end{tabular}
\caption{
Main results across Math, Coding, and SWE.
Values in parentheses denote absolute improvements of EvoTrainer over the corresponding no-RL base model.
The AutoResearch~\citep{karpathy2026autoresearch} row reflects score-only autonomous iteration; the SWE-9B value matches the documented score-dominant subpath ceiling (Section~\ref{subsec:framework-evidence}).
}
\vspace{-0.3cm}
\label{tab:main-results}
\end{table*}

\paragraph{EvoTrainer matches or exceeds prior baselines across all settings.}
Table~\ref{tab:main-results} shows that EvoTrainer achieves the strongest score in every reported column. Against the no-RL base model, gains are statistically significant in every domain (paired Wilcoxon $p<0.001$ throughout; Appendix~\ref{app:significance}). Against the human-engineered RL reference, EvoTrainer delivers statistically significant improvements on SWE-9B ($\Delta=+4.39$, $95\%$ CI $[+2.61,+6.34]$, $p<0.001$) and Math ($\Delta=+2.88$, $p<0.001$), while matching the human reference within the bootstrap CI on SWE-4B and Coding ($p>0.1$ in both); full statistics are reported in Appendix~\ref{app:significance}. EvoTrainer exceeds AutoResearch's score-only iteration in every domain, including Coding where it falls below the no-RL base. We therefore treat the SWE-9B and Math results as the principal quantitative claim, while the SWE-4B and Coding results show that EvoTrainer attains human-engineered performance under autonomous control rather than under expert manual tuning.

\paragraph{Retained recipes diverge across domains.}
RAGEN v2 SNR Filtering is the strongest algorithmic baseline in every column, yet EvoTrainer improves on it throughout. More importantly, the retained EvoTrainer recipes differ substantially across domains---Math toward computation-aware tool augmentation, Coding toward execution-aligned reward shaping with variance-aware filtering, SWE toward a richer behavior-sensitive training pathway---indicating that EvoTrainer adapts interventions to domain-specific bottlenecks rather than selecting one universal template.

\subsection{How Does EvoTrainer Adapt Across Math and Coding?}
\label{subsec:domain-evolution}

\begin{figure}[t]
\centering
\includegraphics[width=\columnwidth]{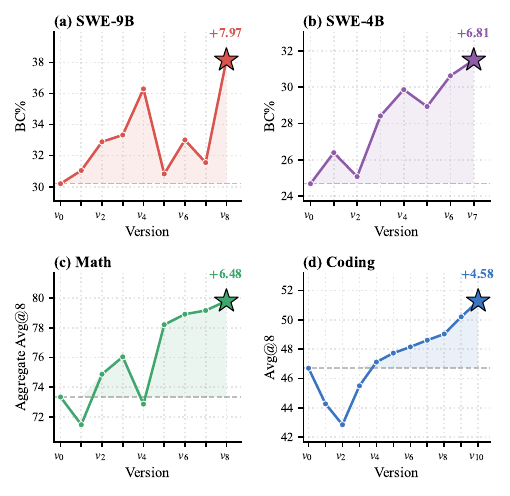}
\caption{
Per-version score trajectories on the promoted path for each training condition.
Stars mark the final retained version.
Dashed lines indicate $v_0$ (no-RL base).
(a)~SWE-9B BC\%; (b)~SWE-4B BC\%; (c)~Math aggregate Avg@8 over AIME 2024 / AIME 2025 / CNMO 2024; (d)~Coding Avg@8.
}
\vspace{-0.3cm}
\label{fig:trajectories}
\end{figure}

We first develop EvoTrainer in the SWE setting, where long-horizon agentic interaction demands the richest diagnostic harness. Math and Coding are then explored with access to the skills and case memory accumulated during SWE. The full SWE version trajectory is reported in Appendix~\ref{app:swe-trajectory}; here we focus on how later domains diverge from one another once that reusable infrastructure is available.

\begin{table*}[t]
\centering
\small
\setlength{\tabcolsep}{4pt}
\begin{tabular}{@{}>{\raggedright\arraybackslash}p{0.19\linewidth}>{\raggedright\arraybackslash}p{0.24\linewidth}>{\raggedright\arraybackslash}p{0.32\linewidth}>{\raggedright\arraybackslash}p{0.20\linewidth}@{}}
\toprule
\textbf{Removed}
& \textbf{Trace}
& \textbf{Outcome}
& \textbf{Implication} \\
\midrule
Richer-than-score evidence
& SWE-9B score-dominant path
& $33.33$ (v3) vs.\ $38.16$ (v8) BC\% ($+4.83$)
& Score-only loop stalls at v3 \\
Harness evolution
& SWE-9B v1 under Git leakage
& 48.80 (invalid) vs.\ 31.04 (true)
& Invalid branch would be promoted \\
Skill library
& Coding v8$\to$v10
& $+1.17$ at v9; $+1.08$ at v10
& Final configuration unreachable \\
\bottomrule
\end{tabular}
\caption{
Component-level counterfactual evidence drawn from the EvoTrainer trajectory.
Each row isolates one EvoTrainer component by comparing the trainer's actual decision with a natural counterfactual already present in the experiment record.
}
\label{tab:framework-evidence}
\end{table*}

In Math, the early diagnosis identifies a response-length bottleneck affecting difficult problems. After length-budget correction and reward-side refinement, residual errors concentrate on computation-heavy cases. A transferred variance-aware filter improves group-level signal but the dominant gap remains computational; the final intervention therefore integrates a Code Interpreter (per-benchmark scores in Table~\ref{tab:main-results}).

Coding follows a different trajectory. The early diagnosis reveals a measurement-side artifact: many zero-reward outputs are format-gate / truncation artifacts rather than semantic coding failures, prompting an output-protocol repair before reward redesign. The trainer then replaces binary correctness with shaped continuous CR based on passed-test ratio, recovering partial execution progress lost under binary scoring. Residual zero-variance groups motivate cross-domain reuse: the trainer retrieves StdGroupFilter from SWE and adapts it, yielding the final shaped CR plus group-variance filtering configuration. Detailed version traces and the transferred skill are in Appendices~\ref{app:domain-evolution} and~\ref{app:stdgroupfilter}.

\subsection{What Happens Without Each Component of EvoTrainer?}
\label{subsec:framework-evidence}

The main results establish performance but not why trainer-level machinery matters. We identify three natural counterfactuals that each isolate one EvoTrainer component without a separate sweep (Table~\ref{tab:framework-evidence}). EvoTrainer retained 33 versions on promoted paths plus 20 negative-evidence candidates; full SWE trajectory and statistical details are in Appendices~\ref{app:swe-trajectory} and~\ref{app:significance}.

\paragraph{Richer diagnostics break the v3 saturation.}
In SWE-9B, an early score-dominant subpath shows only incremental gains ($31.04 \rightarrow 32.89 \rightarrow 33.33$ BC\% across v1--v3) through scalar validation comparison and published-style recipe adaptation before saturating. Once richer diagnostics, backtesting, and harness-guided intervention planning are engaged, the trajectory advances to v4 at 36.30 BC\% and v8 at 38.16 BC\%---a $+4.83$ gain over v3 that score inspection alone did not reach (Appendix~\ref{app:swe-trajectory}).

\paragraph{Harness audit blocks a Git-leak false promotion.}
Under an uncleaned SWE-9B repository state, v1 reaches 48.80 BC\%---a number a score-only loop would promote as a breakthrough. Harness-level inspection detects that the model achieves this by accessing reference patches through Git history commands (\texttt{git show}, \texttt{git log}); after sanitization, the legitimate v1 score is 31.04 BC\%. Without the audit, this invalid branch would have been promoted on misleading scalar evidence (Appendix~\ref{app:swe-trajectory}).

\paragraph{Skill reuse changes the Coding v9 candidate set.}
Coding v9 illustrates how a retained skill alters the trainer's candidate set. Shaped continuous CR at v8 leaves roughly 31\% of rollout groups with near-zero reward variance, matching the SWE v3$\to$v4 regime where StdGroupFilter was validated; the reused filter is therefore a natural candidate. The trainer also evaluates stronger entropy regularization and a lower KL coefficient but rejects both on mechanism rather than score grounds: the former would compound v8's length drift, while the latter would not create within-group variance under a structurally degenerate reward signal. The retained filter drives v9 to $50.21$ Avg@8 ($+1.17$) and seeds the Dual-Level Filter that carries v10 to $51.29$ ($+1.08$); the same skill also improves Math by $+0.96$. Without the skill library, this mechanism-matched candidate would be absent, leaving only rejected local alternatives (Appendix~\ref{app:stdgroupfilter}).

\section{Conclusion}
\label{sec:conclusion}

We introduce EvoTrainer, an autonomous trainer for evidence-conditioned training evolution. Rather than treating autonomous experimentation as repeated recipe edits with scalar score comparison, EvoTrainer treats policy improvement as a cross-version process in which runnable training versions, diagnostic harnesses, and reusable skills evolve together; the trainer agent autonomously runs the diagnostic reasoning loop while humans bootstrap the workspace and gate costly or consequential actions.

Across Math, Coding, and SWE, EvoTrainer consistently improves over no-RL baselines and matches or exceeds the human-engineered RL references, with the largest gain on SWE-9B ($38.16$ vs.\ $33.77$ BC\%). The retained strategies diverge across domains, and component-level counterfactual evidence shows that the gains are not reducible to scalar-score iteration alone. The next step in autonomous model training is to build trainers that learn how to interpret, revise, and improve the training process itself.

\section*{Limitations}

The current realization of EvoTrainer is constrained primarily by compute economics. A complete run consumes approximately $4.0\times10^{8}$ trainer-agent tokens on top of RL training (Appendix~\ref{app:human-budget}); total SWE GPU-hours nonetheless remain below the human-engineered RL reference, suggesting that trainer reasoning substitutes for rather than adds to training-side search. The same constraint motivates a single training seed per version, with stochasticity instead reported through per-task paired bootstrap (Appendix~\ref{app:significance}), following common large-scale LLM-RL practice~\citep{guo2025deepseekr1,yu2025dapo}. Autonomous training with harness evolution further depends on trainer models with strong long-context reasoning and literature-grounded retrieval; our experiments use Claude Sonnet 4.6~\citep{anthropic2026claudesonnet46}. Retained trajectories also span only 7 to 10 versions per domain, leaving behavior over hundreds of versions---where case memory and skill libraries may require active pruning or hierarchical organization---as future work.

\section*{Acknowledgments}

We gratefully acknowledge the support from Tongyi Lab, Alibaba Group, including computational resources, engineering assistance, and helpful discussions during the development of this work. We also thank colleagues and collaborators for their feedback and encouragement throughout the project.




\appendix

\section{Training Objective Details}
\label{app:training-core}

This appendix expands the GRPO-style training core summarized in Section~\ref{subsec:swe-training-core}. The token-level policy ratio is
\[
\rho_{i,t}(\theta) = \frac{\pi_{\theta}(o_{i,t}\mid q, o_{i,<t})}{\pi_{\theta_{\mathrm{old}}}(o_{i,t}\mid q, o_{i,<t})}.
\]
We optimize a clipped policy objective with asymmetric Clip-Higher~\citep{yu2025dapo} bounds (\(\epsilon_\ell = 0.20\), \(\epsilon_u = 0.27\)) and a weak KL term to a fixed reference policy:
\[
\begin{aligned}
\ell_{i,t}(\theta) &= \min\!\left(\rho_{i,t} A_i,\, \mathrm{clip}(\rho_{i,t}, 1 - \epsilon_{\ell}, 1 + \epsilon_u) A_i\right) \\
&\quad - \beta D_{\mathrm{KL}}\!\left(\pi_{\theta}\,\|\,\pi_{\mathrm{ref}}\right)_{i,t}.
\end{aligned}
\]
The full objective averages per-token loss within each trajectory and per-trajectory within each group:
\[
\mathcal{J}(\theta) = \mathbb{E}_{q,\{o_i\}}\!\left[\frac{1}{G}\sum_{i=1}^{G}\frac{1}{|o_i|}\sum_{t=1}^{|o_i|}\ell_{i,t}(\theta)\right].
\]

\section{Additional Domain-Evolution Details}
\label{app:domain-evolution}

This appendix provides additional evidence for the Math and Coding evolution traces summarized in Section~\ref{subsec:domain-evolution}. The main paper reports only the condensed evolution logic needed for the central argument. Here we provide representative version scores, diagnostic indicators, and supporting observations that justify the retained interventions. Intermediate versions that were explored but not retained on the promoted path are omitted for compactness.

\subsection{Math Evolution Trace}

Table~\ref{tab:math-evolution-appendix} summarizes the retained Math trajectory. The evolution follows four stages: identifying a response-length bottleneck, improving reward-side training signal quality, testing a transferred variance-aware filter, and finally introducing a Code Interpreter to address computation-heavy residual errors.

\begin{table*}[t]
\centering
\small
\setlength{\tabcolsep}{5.0pt}
\begin{tabular}{@{}p{0.14\linewidth}p{0.10\linewidth}p{0.23\linewidth}p{0.46\linewidth}@{}}
\toprule
\textbf{Version}
& \textbf{Aggregate Score}
& \textbf{Representative Diagnostic}
& \textbf{Interpretation} \\
\midrule
v0 (Base, no RL)
& 73.33
& truncation rate $\approx 0.18$
& A non-trivial fraction of validation responses reaches the generation limit, motivating length-budget correction. \\
v5
& 78.21
& DGR $\approx 0.21$
& Length unlock, Clip-Higher, and length-aware shaping improve effective group-level learning signal. \\
v7
& 79.17
& DGR $\approx 0.13$
& StdGroupFilter further suppresses residual low-variance groups, but the score gain remains moderate. \\
v8 (EvoTrainer)
& 79.81
& TIR $\approx 0.27$
& Code Interpreter integration addresses computation-heavy residual errors and becomes the retained final intervention. \\
\bottomrule
\end{tabular}
\caption{
Math evolution summary.
The aggregate score is the unweighted mean over AIME 2024, AIME 2025, and CNMO 2024.
DGR denotes Dead Group Ratio and TIR denotes Tool Invocation Rate.
Intermediate versions explored but not retained on the promoted path are omitted.
}
\label{tab:math-evolution-appendix}
\end{table*}

The initial Math diagnosis identifies an upstream truncation bottleneck. Approximately \(18\%\) of validation responses touch the maximum generation length, and difficult competition problems are disproportionately represented among these clipped outputs. This pattern suggests that a subset of failures arises not from an incorrect reasoning strategy, but from reasoning chains that terminate before completion.

After length-budget correction and reward-side refinement, the retained trajectory reaches v5, where the dead-group ratio decreases substantially relative to the earlier regime. However, subsequent failure clustering shows that many remaining errors concentrate on computation-heavy cases, including enumeration-heavy problems, large-number arithmetic, combinatorial expansion, and numerical verification. The failure clustering motivates a shift from reward-only intervention toward external computation support.

The transferred StdGroupFilter in v7 further lowers DGR from approximately \(0.21\) to \(0.13\), indicating cleaner group-level signal. The score gain over v5 remains limited, however, suggesting that the dominant residual bottleneck is not solely low-variance optimization signal. Version v8 therefore integrates a Code Interpreter. In the retained evaluation trace, approximately \(27\%\) of validation samples invoke the tool, and the largest gains concentrate on computation-intensive residual cases. This concentration justifies retaining the tool augmentation as the final intervention: while the aggregate gain over v7 is modest, no reward-side intervention had been able to close the same residual computational gap. The final v8 configuration corresponds to the main-table Math results (Table~\ref{tab:main-results}). The full transfer logic for StdGroupFilter is discussed in Appendix~\ref{app:stdgroupfilter}.

\subsection{Coding Evolution Trace}

Table~\ref{tab:coding-evolution-appendix} summarizes the retained Coding trajectory. The evolution first repairs a measurement-side format artifact, then changes the reward shape to preserve partial-pass information, and finally reuses a SWE-derived filtering skill to suppress residual low-information groups. Intermediate versions that were explored but not retained on the promoted path are omitted for compactness.

\begin{table*}[t]
\centering
\small
\setlength{\tabcolsep}{5.0pt}
\begin{tabular}{@{}p{0.14\linewidth}p{0.08\linewidth}p{0.25\linewidth}p{0.46\linewidth}@{}}
\toprule
\textbf{Version}
& \textbf{Avg@8}
& \textbf{Representative Diagnostic}
& \textbf{Interpretation} \\
\midrule
v0 (Base, no RL)
& 46.71
& Fmt0 $\approx 0.50$
& A large share of zero-reward cases arises from format-gate and truncation interaction. \\
v8
& 49.04
& MidBand $\approx 0.34$, DGR $\approx 0.31$
& Shaped continuous CR recovers partial-pass execution signal and improves group-level contrast. \\
v9
& 50.21
& DGR $\approx 0.21$
& Transferred StdGroupFilter removes residual zero-variance groups after reward shaping. \\
v10 (EvoTrainer)
& 51.29
& DGR $\approx 0.18$, ADiv $\approx 0.63$
& Dual-Level Filter supports broader algorithmic exploration. \\
\bottomrule
\end{tabular}
\caption{
Coding evolution summary.
Fmt0 denotes format-zero rate, MidBand denotes the fraction of samples with \(0 < \mathrm{pass\ ratio} < 1\), DGR denotes Dead Group Ratio, and ADiv denotes algorithm-selection diversity.
Intermediate versions explored but not retained on the promoted path are omitted.
}
\label{tab:coding-evolution-appendix}
\end{table*}

For compactness, the retained trace table omits versions v1 through v7; their main findings are summarized in prose as two intermediate milestones: v3 closes the dominant format-gate false-negative channels, while v7 consolidates budget and regularization calibration before stable shaped-CR training begins at v8.

The initial Coding diagnosis reveals that part of the apparent model failure is actually a measurement-side artifact. Under the early format gate, many generations are assigned zero reward because output truncation prevents the required reasoning/code structure from being closed properly. This motivates an output-protocol repair before reward redesign.

After protocol cleanup, the dominant bottleneck shifts toward sparse binary feedback. A substantial fraction of failing programs passes some, but not all, test cases. Binary correctness collapses these partial-pass trajectories into the same reward value as completely failing programs. Version v8 therefore replaces binary CR with shaped continuous CR based on the passed-test ratio. The resulting MidBand mass rises to approximately \(0.34\), and DGR decreases to approximately \(0.31\), indicating that more rollout groups now contain useful within-group reward variation.

Residual zero-variance groups remain even after reward shaping. This motivates v9, which transfers the StdGroupFilter from SWE into Coding. DGR further decreases to approximately \(0.21\). The final v10 configuration extends StdGroupFilter into a Dual-Level Filter, reaching 51.29 Avg@8 with DGR \(\approx 0.18\) and ADiv \(\approx 0.63\). The Dual-Level Filter (adopted only in Coding, where multi-test rollouts produce nontrivial rates of anomalous trajectories) combines two complementary filtering stages within a single training step. (i) A \emph{trajectory-level} filter marks individual rollouts with anomalous status---truncation, max-length exceedance, or judge error---and excludes them from advantage computation; trajectory-level stabilization in agentic RL has been studied under StarPO-S \citep{wang2025ragen}. (ii) The \emph{group-level} StdGroupFilter (Appendix~\ref{app:stdgroupfilter}) discards low-variance rollout groups via an EMA-tracked top-$p$ threshold over within-group reward standard deviation, building on the dynamic-sampling principle of DAPO \citep{yu2025dapo} and the SNR-aware variance filtering of RAGEN-2 \citep{wang2026ragen2}. The retained trace therefore supports the interpretation that Coding improvement comes from a sequence of measurement repair, signal revision, and reusable skill transfer rather than from a single static recipe. The full transfer analysis is provided in Appendix~\ref{app:stdgroupfilter}.

\subsection{Format-Gate Repair in Coding}

The Coding format-gate repair is a concrete example of harness evolution uncovering a measurement artifact before model-side intervention. In the affected evaluation trace, validation checkpoints at steps \(0\), \(25\), and \(375\) together contain 700 sampled responses. The response-token distribution is heavily concentrated at the former generation limit: the \(p50\), \(p90\), and \(p99\) token lengths all reach 20,480 tokens. Moreover, \(62.1\%\) of samples contain at least 18,000 tokens, and \(56.1\%\) fail to close the reasoning block with \texttt{</think>}.

Under a stricter format-fail criterion that requires both missing \texttt{</think>} and missing closed code block, 352 samples qualify; all of them satisfy \(n_{\mathrm{tok}}\geq18{,}000\). This concentration shows that the zero-reward format failures are not randomly distributed semantic mistakes; they are tightly coupled to truncation under the previous output budget. The subsequent repair increases \texttt{max\_new\_tokens} from 20,480 to at least 32,768, allowing the format gate to operate on more complete outputs. This cleanup precedes reward redesign and prevents later training decisions from being driven by an avoidable measurement artifact.

\subsection{Diagnostic Metric Glossary}

Table~\ref{tab:metric-glossary} summarizes the diagnostic indicators that appear in the Math and Coding auxiliary analyses.

\begin{table*}[t]
\centering
\small
\setlength{\tabcolsep}{6.0pt}
\begin{tabular}{@{}p{0.10\linewidth}p{0.53\linewidth}p{0.30\linewidth}@{}}
\toprule
\textbf{Metric}
& \textbf{Definition}
& \textbf{Preferred Direction} \\
\midrule
DGR
& Dead-group ratio: fraction of rollout groups with near-zero within-group reward variance
& Lower \\
TIR
& Tool Invocation Rate: fraction of samples invoking \texttt{python\_execute}
& Task-dependent; useful when tied to correctness \\
Fmt0
& Format-zero rate: fraction of samples forced to zero reward by formatting failure
& Lower \\
MidBand
& Fraction of samples with \(0 < \mathrm{pass\ ratio} < 1\)
& Higher when partial execution progress is useful \\
ADiv
& Algorithm-selection diversity entropy within sampled groups
& Higher when broader exploration is desired \\
\bottomrule
\end{tabular}
\caption{
Glossary of diagnostic indicators used in the Math and Coding auxiliary analyses.
}
\label{tab:metric-glossary}
\end{table*}

\section{Cross-Domain Reuse of StdGroupFilter}
\label{app:stdgroupfilter}

The StdGroupFilter provides a concrete example of reusable skill accumulation in EvoTrainer, as introduced in Section~\ref{subsec:memory} and applied across domains in Section~\ref{subsec:domain-evolution}. Rather than storing only natural-language observations, the trainer preserves validated operational mechanisms that can be retrieved and adapted in later domains. StdGroupFilter originates in SWE, where the trainer diagnoses low-information rollout groups with insufficient reward variance. Because the mechanism operates on group-level statistics rather than SWE-specific reward semantics, it later becomes reusable in both Math and Coding.

\subsection{Origin in SWE}

In the SWE analysis pipeline, the trainer identifies a recurring failure mode: rollout groups with insufficient reward variance contribute little useful relative-learning signal while still consuming training compute. Group-level filtering by reward variance has been studied under Dynamic Sampling in DAPO \citep{yu2025dapo} and SNR-Aware Filtering in RAGEN-2 \citep{wang2026ragen2}; building on these ideas, the trainer instantiates a filtering utility that tracks the reward-standard-deviation distribution via an EMA-tracked top-$p$ threshold, retaining informative groups more flexibly than a rigid fixed-threshold rule.

The retained configuration uses:

\begin{itemize}
    \item \texttt{group\_filter\_mode: top\_p}
    \item \texttt{group\_keep\_ratio: 0.9}
    \item \texttt{group\_min\_keep\_ratio: 0.5}
    \item \texttt{group\_ema\_decay: 0.1}
    \item \texttt{group\_min\_std\_threshold: 0.01}
\end{itemize}

The filter operates on group-level reward statistics rather than on SWE-specific components such as CR, SBE, ETT, or IF. This abstraction is central to its later cross-domain transfer. Table~\ref{tab:stdgroupfilter-transfer} summarizes these cross-domain transfers.

\begin{table*}[t]
\centering
\small
\setlength{\tabcolsep}{5.2pt}
\begin{tabular}{@{}p{0.08\linewidth}p{0.17\linewidth}p{0.34\linewidth}p{0.34\linewidth}@{}}
\toprule
\textbf{Domain}
& \textbf{Adoption Stage}
& \textbf{Observed Change}
& \textbf{Role in the Trace} \\
\midrule
SWE
& Initial development
& Introduced to suppress low-information reward groups
& Skill origin \\
Math
& v5 $\rightarrow$ v7
& Aggregate score \(78.21 \rightarrow 79.17\)
& Cross-domain reuse with moderate gain \\
Coding
& v8 $\rightarrow$ v9
& Avg@8 \(49.04 \rightarrow 50.21\)
& Cross-domain reuse after shaped CR \\
Coding
& v9 $\rightarrow$ v10
& Avg@8 \(50.21 \rightarrow 51.29\)
& Retained inside the final Dual-Level Filter \\
\bottomrule
\end{tabular}
\caption{
Cross-domain reuse of StdGroupFilter.
The same group-level filtering skill is developed in SWE, later retrieved in Math and Coding, and ultimately retained as part of the final Coding configuration.
}
\label{tab:stdgroupfilter-transfer}
\end{table*}

\subsection{Transfer to Math}

In Math, the trainer recognizes that residual zero-variance groups after reward-side improvements are structurally similar to the earlier SWE diagnosis. This transfer is appropriate because the filter depends on within-group reward dispersion rather than on task-specific reward semantics. The corresponding version-level changes are reported in Table~\ref{tab:math-evolution-appendix}.

\subsection{Transfer to Coding}

The Coding trace later exhibits a related optimization pathology. Even after shaped continuous CR recovers partial-pass signal, many sampled groups remain effectively low-information: some groups fail uniformly, some pass the same fixed subset of easy tests, and others are already uniformly solved. These cases motivate a group-level filter that can complement, rather than replace, component-level reward shaping. The trainer retrieves the previously validated StdGroupFilter from the reusable skill library and adapts it to the Coding pipeline. The corresponding version-level context is reported in Table~\ref{tab:coding-evolution-appendix}.

\section{Additional SWE Framework and Trajectory Diagnostics}
\label{app:swe-trajectory}

This appendix provides supplementary evidence for Section~\ref{subsec:framework-evidence}. We first expand the score-dominant SWE-9B path discussed in the main text, then report the Git-leak audit, the IF-Judge dead-group backtest, and additional behavior-level degeneration cases that motivate richer harness diagnostics.

\subsection{Score-Dominant SWE-9B Path versus Full EvoTrainer}

Table~\ref{tab:score-dominant-trace} expands the score-dominant SWE-9B path summarized in the main paper. The early path relies on scalar BC\% comparison and published-style recipe adaptation. It improves over the no-RL baseline but saturates before reaching the stronger EvoTrainer regimes.

\begin{table*}[t]
\centering
\small
\setlength{\tabcolsep}{5.2pt}
\begin{tabular}{llll}
\toprule
\textbf{Version}
& \textbf{Training Core}
& \textbf{Main Intervention}
& \textbf{SWE-9B BC\%} \\
\midrule
v0 (Base, no RL)
& No RL
& --
& 30.19 \\
v1
& GSPO
& Flat correctness reward
& 31.04 \\
v2
& GRPO
& RAGEN-style filtering
& 32.89 \\
v3
& GRPO + Clip-Higher
& Continuous CR + Top-K filtering
& 33.33 \\
v4
& GRPO + Clip-Higher
& Binary CR + SBE + ETT + EMA filtering
& 36.30 \\
v8
& GRPO + Clip-Higher
& v4 recipe + IF LLM Judge
& 38.16 \\
\bottomrule
\end{tabular}
\caption{
Detailed SWE-9B path underlying the framework-level contrast in the main text.
The score-dominant early path saturates at 33.33 BC\%, whereas the full EvoTrainer trajectory reaches 38.16 BC\%.
}
\label{tab:score-dominant-trace}
\end{table*}

The training core was adjusted in two early steps: GSPO at v1 (flat correctness reward, +0.85 BC\% over no-RL) was switched to GRPO at v2, with Clip-Higher~\citep{yu2025dapo} added from v3. We do not present the v1$\rightarrow$v2 step as a controlled algorithm comparison since RAGEN-style filtering was added in the same iteration. The substantive intervention occurs at v3$\rightarrow$v4: by v3, three reward and filtering refinements had saturated at 33.33 BC\%, prompting a coupled reward redesign---procedural rewards (SBE, ETT), binary correctness, and EMA-based group filtering on top of GRPO. After v4 reached 36.30 BC\%, the trainer explored more aggressive variants on top of this branch at v5--v7, each of which regressed below v4 on validation; the trainer therefore reverted to the v4 branch as the baseline for further iteration. v8 then extends v4 with an instruction-following judge, after dead-group analysis and reward backtesting indicate that the existing reward structure leaves substantial training variance unused.

\subsection{Harness Audit Prevents False Promotion under Git Leakage}

This audit case illustrates why scalar validation scores alone are insufficient for autonomous training decisions. In an early SWE-9B v1 environment, the model could access Git history commands such as \texttt{git show} and \texttt{git log}. This unintentionally exposed reference-patch information and produced an apparent validation score of 48.80 BC\%. A score-only training loop would have treated this branch as a decisive improvement and promoted it as the best version.

The evolving SWE harness identifies the anomaly through behavior-level inspection of tool usage and repository interactions. In particular, the audit detects anomalous Git-command usage together with a late jump in response length. After the environment is sanitized by removing access to the leaked Git history, the legitimate v1 performance is 31.04 BC\%, which is the value used throughout the main experiments and the score reported in the retained SWE evolution path. Table~\ref{tab:git-leak-audit} reports the contaminated and sanitized evaluation results.

\begin{table*}[t]
\centering
\small
\setlength{\tabcolsep}{6.0pt}
\begin{tabular}{lcc}
\toprule
\textbf{Evaluation Condition}
& \textbf{Observed SWE-9B v1 BC\%}
& \textbf{Interpretation} \\
\midrule
Contaminated repository state
& 48.80
& Invalid apparent gain caused by Git-history leakage \\
Sanitized repository state
& 31.04
& Legitimate v1 performance used in the paper \\
\bottomrule
\end{tabular}
\caption{
Git-leak audit for SWE-9B v1.
The contaminated score would falsely dominate the retained trajectory under scalar-score-only selection, while harness inspection prevents invalid promotion.
}
\label{tab:git-leak-audit}
\end{table*}

This case is central to our distinction between score-driven iteration and trainer-level diagnosis. The harness changes the branch-selection decision itself by rejecting an apparently high-scoring but invalid trajectory.

\subsection{Dead-Group Rescue with the IF LLM Judge}

Version v4 already reaches 36.30 BC\%, but trainer-side diagnostics show that the dead-group ratio remains high. Approximately half of the rollout groups still produce no useful within-group reward variation, which limits the effectiveness of group-relative optimization.

To test whether an additional reward dimension can recover useful variance, the trainer introduces an instruction-following signal and performs retroactive reward backtesting on historical rollouts from the promoted SWE-9B branch. These rollouts are re-scored after augmenting the original correctness signal with a \(0.1\)-weighted instruction-following term, and within-group reward variance is then recomputed. Among groups that are dead under the original correctness signal, \(45\%\) regain non-zero variance after adding the instruction-following term. Table~\ref{tab:if-judge-dead-groups} summarizes the resulting dead-group reduction.

\begin{table*}[t]
\centering
\small
\setlength{\tabcolsep}{5.4pt}
\begin{tabular}{lccc}
\toprule
\textbf{Version}
& \textbf{Reward Design}
& \textbf{Dead Group Ratio}
& \textbf{BC\%} \\
\midrule
v1
& Flat correctness reward
& 55.0\%
& 31.04 \\
v4
& Binary CR + SBE + ETT + EMA filtering
& 50.0\%
& 36.30 \\
v8
& v4 recipe + IF LLM Judge
& 27.5\%
& 38.16 \\
\bottomrule
\end{tabular}
\caption{
Dead-group reduction across selected SWE-9B versions.
The IF LLM Judge is introduced after backtesting indicates that it restores useful variance in otherwise low-information groups.
}
\label{tab:if-judge-dead-groups}
\end{table*}

This diagnostic chain clarifies why v8 is more than a generic additive reward term. Its introduction follows a mechanism-level analysis of residual dead groups and is validated through offline reward backtesting before becoming part of the retained training recipe.

\subsection{Echo Trap in SWE-4B}

The preceding subsections focus on SWE-9B. We next report complementary SWE-4B analyses that expose additional behavior-level failure modes observed during training. The Echo Trap refers to a trajectory-degeneration pattern in which turns become excessively long and repetitive while validation score may still appear temporarily competitive. Table~\ref{tab:echo-trap} summarizes a representative SWE-4B branch.

\begin{table*}[t]
\centering
\small
\setlength{\tabcolsep}{6.0pt}
\begin{tabular}{cccccc}
\toprule
\textbf{Step}
& \textbf{BC\%}
& \textbf{Turns}
& \textbf{VLong \(>70\)}
& \textbf{Filtered}
& \textbf{Shaped Reward} \\
\midrule
0   & --    & 35.7 & 7.8\%  & 23  & 0.59 \\
75  & 29.71 & 36--37 & -- & -- & -- \\
125 & 33.28 & 76.3 & 60.7\% & 194 & 0.45 \\
\bottomrule
\end{tabular}
\caption{
Representative SWE-4B Echo Trap degeneration.
VLong $>70$ denotes the fraction of trajectories exceeding 70 turns.
The validation score rises between steps 75 and 125, but behavior-level diagnostics show severe trajectory elongation and a sharp increase in filtered degenerate rollouts.
Step-75 behavior-level breakdowns were not yet standardized in the original trace; BC\% and average turns are available at that checkpoint, while the remaining indicators were retained only in aggregate form.
}
\label{tab:echo-trap}
\end{table*}

Table~\ref{tab:echo-trap} makes the contrast explicit: BC\% rises from 29.71 to 33.28, while average turn count more than doubles (36--37 to 76.3) and the filtered degenerate-trajectory count rises eightfold (23 to 194). Headline validation improvement alone is insufficient to certify a branch as healthy.

The associated behavior-level filter distinguishes three degeneration modes:

\begin{enumerate}
    \item Abandonment: trajectories with at most 3 turns and no edit action;
    \item Idle failure: trajectories with no edit action and zero correctness reward;
    \item Echo Trap: trajectories exceeding 60 turns while still receiving zero correctness reward.
\end{enumerate}

The filtered count grows from 23 in the early regime to 194 in the collapse regime, consistent with a major shift toward degenerate behavior rather than ordinary score variance. After rejecting this echo-trapped branch through behavior-level audit, the retained SWE-4B configuration reported in Table~\ref{tab:main-results} comes from an alternative non-collapsing branch.

\subsection{Multiplicative Efficiency Factor Collapse (SWE-4B Branches)}

All version labels (v9, v10, v11) in this subsection refer to SWE-4B branches investigated after the Echo Trap diagnosis. Following that diagnosis, a three-way branch test investigates how the system should respond after long-trajectory degeneration is detected. The multiplicative efficiency factor takes the form
\[
\mathrm{efficiency}
=
1.0 - 0.3 \times \frac{\mathrm{num\_turns}}{\mathrm{max\_turns}},
\]
with \(\mathrm{efficiency}\in[0.7,1.0]\). In v9, this factor multiplies a staircase correctness reward; in v11, it multiplies continuous CR. Version v10 serves as a control branch with continuous CR but without the multiplicative efficiency factor.

\paragraph{v9: Efficiency Only.}
Table~\ref{tab:v9-efficiency-collapse} shows that the branch initially reduces average turn count, but then collapses into near-trivial one-turn behavior. BC\% falls to approximately zero, dead groups approach \(100\%\), and diversity disappears.

\begin{table*}[t]
\centering
\small
\setlength{\tabcolsep}{5.2pt}
\begin{tabular}{ccccccc}
\toprule
\textbf{Step}
& \textbf{BC\%}
& \textbf{CR}
& \textbf{Dead\%}
& \textbf{Diversity\%}
& \textbf{AvgTurns}
& \textbf{P90Turns} \\
\midrule
0   & 21.3 & 0.537 & 19.5  & 80.5 & 36.7 & 69 \\
50  & 18.0 & 0.572 & 58.4  & 41.6 & 7.3  & 15 \\
100 & 9.3  & 0.531 & 68.8  & 31.2 & 3.0  & 5 \\
150 & 0.0  & 0.495 & 100.0 & 0.0  & 1.0  & 1 \\
200 & 0.0  & 0.493 & 97.4  & 2.6  & 1.0  & 1 \\
250 & 0.2  & 0.496 & 98.7  & 1.3  & 1.0  & 1 \\
\bottomrule
\end{tabular}
\caption{
Collapse under multiplicative efficiency shaping in SWE v9.
}
\label{tab:v9-efficiency-collapse}
\end{table*}

\paragraph{v11: Continuous CR plus Efficiency.}
Table~\ref{tab:v11-efficiency-collapse} shows the same failure pattern even when continuous CR replaces the staircase reward. The branch again converges toward one-turn behavior and near-total dead-group saturation.

\begin{table*}[t]
\centering
\small
\setlength{\tabcolsep}{5.2pt}
\begin{tabular}{ccccccc}
\toprule
\textbf{Step}
& \textbf{BC\%}
& \textbf{CR}
& \textbf{Dead\%}
& \textbf{Diversity\%}
& \textbf{AvgTurns}
& \textbf{P90Turns} \\
\midrule
0   & 24.7 & 0.834 & 27.3 & 72.7 & 36.5 & 65 \\
50  & 19.2 & 0.892 & 37.7 & 62.3 & 6.2  & 13 \\
100 & 17.7 & 0.892 & 39.0 & 61.0 & 4.2  & 7 \\
150 & 0.5  & 0.911 & 93.5 & 6.5  & 1.2  & 2 \\
175 & 0.0  & 0.910 & 97.4 & 2.6  & 1.0  & 1 \\
\bottomrule
\end{tabular}
\caption{
Collapse under continuous CR combined with multiplicative efficiency shaping in SWE v11.
}
\label{tab:v11-efficiency-collapse}
\end{table*}

\paragraph{v10: Continuous CR without Efficiency.}
Table~\ref{tab:v10-control} provides the counterfactual branch. Continuous CR without the multiplicative efficiency term remains viable over hundreds of steps and does not collapse into one-turn behavior.

\begin{table*}[t]
\centering
\small
\setlength{\tabcolsep}{5.2pt}
\begin{tabular}{ccccccc}
\toprule
\textbf{Step}
& \textbf{BC\%}
& \textbf{CR}
& \textbf{Dead\%}
& \textbf{Diversity\%}
& \textbf{AvgTurns}
& \textbf{P90Turns} \\
\midrule
0   & 23.5 & 0.821 & 20.8 & 79.2 & 36.8 & 66 \\
50  & 27.1 & 0.798 & 31.2 & 68.8 & 41.2 & 77 \\
100 & 26.9 & 0.862 & 28.6 & 71.4 & 44.3 & 76 \\
150 & 29.5 & 0.860 & 26.0 & 74.0 & 44.2 & 76 \\
200 & 26.6 & 0.889 & 42.9 & 57.1 & 36.6 & 62 \\
250 & 27.9 & 0.887 & 37.7 & 62.3 & 43.4 & 72 \\
350 & 26.5 & 0.894 & 41.6 & 58.4 & 39.0 & 68 \\
\bottomrule
\end{tabular}
\caption{
Control branch with continuous CR but without the multiplicative efficiency factor.
}
\label{tab:v10-control}
\end{table*}

The v9/v11 failures, together with the v10 control, show that a seemingly modest reward multiplier can produce catastrophic behavioral incentives in agentic RL. EvoTrainer retains these branches as reusable negative evidence rather than treating them as disposable failed runs.

\section{Search-Budget and Compute Accounting for Human Baselines}
\label{app:human-budget}

This appendix documents the fairness conditions and search-budget accounting for the human-engineered RL references used in the main results. While the comparison is not perfectly compute-matched, it is transparent enough to show that the EvoTrainer gains are not explained by privileged access to data, infrastructure, or weaker human baselines.

\subsection{Shared Experimental Conditions}

The human and EvoTrainer conditions share the following properties:

\begin{itemize}
    \item Both begin from the same project codebase and official training scripts.
    \item Both use the same task assets, data splits, model families, hardware environment, and evaluation protocol.
    \item All reported numbers use the same validation setup and the same seed convention.
    \item The paper reports the strongest human-engineered RL configuration observed under this shared stack, rather than an average or first attempt.
\end{itemize}

Under these conditions, the main difference is the decision-making process used to select subsequent interventions: human RL engineering versus the trainer-agent workflow in EvoTrainer.

\subsection{SWE Search Budget}

Table~\ref{tab:human-trace-budget} summarizes the approximate search-budget comparison for SWE. The human condition retains more total versions, runs more total training steps, and uses more total GPU-hours, while EvoTrainer achieves stronger final scores on both SWE-4B and SWE-9B.

\begin{table*}[t]
\centering
\small
\setlength{\tabcolsep}{5.2pt}
\begin{tabular}{lccc}
\toprule
\textbf{Quantity}
& \textbf{EvoTrainer}
& \textbf{Human-Engineered RL}
& \textbf{Comparison} \\
\midrule
SWE retained versions
& 15 (SWE-9B: 8 / SWE-4B: 7)
& \(\sim\)23 (combined estimate)
& Human covers more \\
SWE total training steps
& \(\sim 3{,}000\)
& \(\sim 4{,}600\)
& Human \(\sim 1.53\times\) steps \\
SWE total GPU-hours
& \(\sim 92{,}800\)
& \(\sim 140{,}000\)
& Human \(\sim 1.5\times\) GPU-hours \\
SWE-4B final BC\%
& 31.49
& 31.17
& EvoTrainer \(+0.32\) \\
SWE-9B final BC\%
& 38.16
& 33.77
& EvoTrainer \(+4.39\) \\
\bottomrule
\end{tabular}
\caption{
Approximate SWE search-budget comparison between EvoTrainer and the human-engineered RL line.
EvoTrainer retains SWE-9B and SWE-4B versions on independent trajectories, while the human-engineered RL line tracks one combined version-count estimate across both model sizes.
}
\label{tab:human-trace-budget}
\end{table*}

The comparison indicates that EvoTrainer does not obtain its SWE gains by simply consuming a larger raw training budget. The human line uses more total steps and more total GPU-hours, yet EvoTrainer achieves stronger retained models, particularly in SWE-9B.

\subsection{Math and Coding Budget Alignment}

Math and Coding are compared under aligned version-level budgets. In Math, EvoTrainer and the human-engineered RL line each retain 8 major versions on the promoted path. In Coding, each retains 10 major versions. Because the retained version counts and training setup are aligned within each domain, total training time is also broadly comparable. Table~\ref{tab:math-coding-budget} summarizes the version-level alignment.

\begin{table*}[t]
\centering
\small
\setlength{\tabcolsep}{5.6pt}
\begin{tabular}{lccc}
\toprule
\textbf{Domain}
& \textbf{EvoTrainer Retained}
& \textbf{Human Retained}
& \textbf{Budget Interpretation} \\
\midrule
Math
& 8
& 8
& Retained-version budget aligned \\
Coding
& 10
& 10
& Retained-version budget aligned \\
\bottomrule
\end{tabular}
\caption{
Retained version-level budget alignment for Math and Coding.
}
\label{tab:math-coding-budget}
\end{table*}

Under these aligned iteration budgets, EvoTrainer reaches 84.17 / 73.33 / 81.94 on AIME 2024 / AIME 2025 / CNMO 2024 (vs.\ 80.83 / 71.67 / 77.78 for the human reference) and 51.29 Coding Avg@8 (vs.\ 50.71).

These comparisons support the interpretation that the advantage of EvoTrainer is not restricted to a single unusually favorable search budget, but appears under both lower-budget SWE search and aligned-budget single-turn domains.

\subsection{Trainer-Agent Inference Usage}

In addition to RL training compute, the full EvoTrainer development workflow consumed approximately \(4.0\times10^8\) trainer-agent tokens for diagnosis, evidence retrieval, backtesting, and intervention planning. We report this separately from GPU-hour accounting because it reflects LLM-mediated analysis and planning rather than policy-training compute.

\section{Head-to-Head Significance Analysis}
\label{app:significance}

This appendix expands the significance numbers cited in Section~\ref{subsec:main-results}. We report three families of paired statistical comparisons: head-to-head against the human-engineered RL best (the headline comparison), head-to-head against the no-RL base model (to confirm that improvements over the seed are not domain-specific artefacts), and within-trajectory comparisons grounding the component-level counterfactual evidence in Section~\ref{subsec:framework-evidence}.

\paragraph{Protocol.}
Each per-task observation is the per-prompt Avg@8 score over $8$ independent rollouts; observations therefore live on the discrete grid $\{0/8,\,1/8,\,\ldots,\,8/8\}$. For two methods $A$ and $B$ we compute the per-task paired difference $d_i = s^A_i - s^B_i$ and report (i) the mean difference $\Delta = \tfrac{1}{n}\sum_i d_i$, (ii) a $95\%$ confidence interval from a paired bootstrap with $B=10{,}000$ resamples drawn at the task level \citep{agarwal2021precipice,colas2018howmany}, and (iii) a two-sided Wilcoxon signed-rank $p$-value \citep{henderson2018matters}. For Math we use a single stratified bootstrap that draws within each of the three sub-benchmarks (AIME 2024, AIME 2025, CNMO 2024) and aggregates the per-task differences across all $n=78$ problems. The Math aggregate scores reported in Table~\ref{tab:sig-vs-human} are therefore problem-weighted means over the 78 paired observations and may differ by a small amount from the unweighted sub-benchmark means reported in Appendix~\ref{app:domain-evolution}. This protocol substitutes evaluation-time stochasticity for the training-seed stochasticity that is infeasible to repeat at our compute scale, following the dominant LLM-RL reporting practice \citep{guo2025deepseekr1,yu2025dapo}.

\begin{table*}[t]
\centering
\small
\setlength{\tabcolsep}{4pt}
\begin{tabular}{lcccc}
\toprule
\textbf{Domain} & $n$ & \textbf{EvoTrainer} / \textbf{Human-RL} & $\Delta$ \textbf{(95\% CI)} & $p$ \\
\midrule
SWE-9B  & 77  & $38.16$ / $33.77$ & $+4.39\ [+2.61,+6.34]$ & $<\!0.001$ \\
SWE-4B  & 77  & $31.49$ / $31.17$ & $+0.32\ [-1.46,+2.11]$ & $0.797$ \\
Math    & 78  & $79.49$ / $76.60$ & $+2.88\ [+1.12,+4.81]$ & $<\!0.001$ \\
Coding  & 175 & $51.29$ / $50.71$ & $+0.57\ [-0.18,+1.34]$ & $0.142$ \\
\bottomrule
\end{tabular}
\caption{
EvoTrainer vs.\ Human-engineered RL.
Per-task paired bootstrap ($B=10{,}000$) and two-sided Wilcoxon signed-rank tests on per-task Avg@8.
The Math row aggregates AIME 2024 ($n{=}30$), AIME 2025 ($n{=}30$) and CNMO 2024 ($n{=}18$) under a stratified bootstrap.
}
\label{tab:sig-vs-human}
\end{table*}

\begin{table*}[t]
\centering
\small
\setlength{\tabcolsep}{4pt}
\begin{tabular}{lccc}
\toprule
\textbf{Domain} & $n$ & $\Delta$ \textbf{(95\% CI)} & $p$ \\
\midrule
SWE-9B  & 77  & $+7.95\ [+5.52,+10.39]$ & $<\!0.001$ \\
SWE-4B  & 77  & $+6.82\ [+4.55,+9.09]$  & $<\!0.001$ \\
Math    & 78  & $+6.41\ [+4.17,+8.81]$  & $<\!0.001$ \\
Coding  & 175 & $+4.57\ [+3.21,+5.93]$  & $<\!0.001$ \\
\bottomrule
\end{tabular}
\caption{
EvoTrainer vs.\ no-RL base model.
Same protocol as Table~\ref{tab:sig-vs-human}.
}
\label{tab:sig-vs-base}
\end{table*}

\paragraph{Findings.}
The headline gain on SWE-9B and the aggregate Math improvement (Table~\ref{tab:sig-vs-human}) are both statistically significant at the $p<0.001$ level. The smaller margins on SWE-4B and Coding sit within the bootstrap confidence interval and are not statistically distinguishable from the human-engineered RL reference; we therefore frame these settings as EvoTrainer matching the human reference rather than exceeding it. The EvoTrainer-vs-Base comparison (Table~\ref{tab:sig-vs-base}) shows that EvoTrainer delivers significant improvements over the no-RL starting point in every domain, including the two settings where the gap to the human-engineered RL reference is statistically inconclusive. This rules out the alternative reading that the autonomous training loop fails to add value in those domains: in SWE-4B and Coding, EvoTrainer reproduces the gains achieved by hand-engineered RL without expert manual tuning, rather than failing to improve over the base.

\paragraph{Within-Trajectory Paired Comparisons.}
For the component-level counterfactual evidence in Section~\ref{subsec:framework-evidence}, Table~\ref{tab:sig-within-trajectory} reports within-trajectory paired statistics under the same per-task paired bootstrap and Wilcoxon protocol as Table~\ref{tab:sig-vs-human}.

\begin{table*}[t]
\centering
\small
\setlength{\tabcolsep}{4pt}
\begin{tabular}{lccc}
\toprule
\textbf{Comparison} & $n$ & $\Delta$ \textbf{(95\% CI)} & $p$ \\
\midrule
SWE-9B v3 $\to$ v8  & 77  & $+4.83\ [+2.84,+6.95]$  & $<\!0.001$ \\
Coding v8 $\to$ v9  & 175 & $+1.17\ [+0.42,+1.94]$  & $0.003$ \\
Coding v9 $\to$ v10 & 175 & $+1.08\ [+0.36,+1.83]$  & $0.005$ \\
\bottomrule
\end{tabular}
\caption{
Within-trajectory paired comparisons referenced in Section~\ref{subsec:framework-evidence}.
}
\label{tab:sig-within-trajectory}
\end{table*}

\end{document}